# RECENT ADVANCEMENTS IN THE FIELD OF DEEPFAKE DETECTION

Natalie Krueger[1], Dr. Mounika Vanamala[1], Dr. Rushit Dave[2]

[1]Department of Computer Science, University of Wisconsin-Eau Claire, Eau Claire, Wisconsin, USA
kruegena4954@uwec.edu
vanamalm@uwec.edu

[2]Department of Computer Information Science Minnesota State University, Mankato, Mankato, MN, USA
rushit.dave@mnsu.edu

## ABSTRACT

*A deepfake is a photo or video of a person whose image has been digitally altered or partially replaced with an image of someone else. Deepfakes have the potential to cause a variety of problems and are often used maliciously. A common usage is altering videos of prominent political figures and celebrities. These deepfakes can portray them making offensive, problematic, and/or untrue statements. Current deepfakes can be very realistic, and when used in this way, can spread panic and even influence elections and political opinions. There are many deepfake detection strategies currently in use but finding the most comprehensive and universal method is critical. So, in this survey we will address the problems of malicious deepfake creation and the lack of universal deepfake detection methods. Our objective is to survey and analyze a variety of current methods and advances in the field of deepfake detection.*

## 1. INTRODUCTION

Deepfakes are a big issue in the world today. The main reason for this is that deepfakes can be created with malicious intent. Most malicious deepfakes feature well-known figures in society such as politicians or celebrities. Thus causes a danger because deepfakes featuring these people can make it appear as though they are saying something controversial. A person who does not understand deepfakes could be influenced by this. Even more concerning is that deepfakes are easier to produce than ever, and even someone with very little knowledge of technology can use premade methods to create them.

We have chosen to first analyze neural networks as a method of deepfake detection because they are the most common and most researched method. Many studies have found promising results with the use of neural networks [1-28], so surveying these results is essential to understanding deepfake detection as a whole and determining the most accurate methods. We will also be surveying methods that utilize tools other than neural networks [29-32]. It is important to consider newer and more innovative methods when searching for the best results, and many researchers have branched out from neural networks both to evaluate new methods entirely or to compare the results of new methods with neural network-based methods. Finally, we will survey studies that focus on the use of deepfake datasets in deepfake detection [33-35]. The dataset(s) used in training and testing of neural networks and other methods can have a substantial impact on results. We will consider methods that focus on this impact in our survey and analyze their results.

## 2. BACKGROUND

### 2.1. Acknowledgements

A deepfake is a digitally altered photo, video, or audio that contains a person. In a deepfake, the image or voice of the person is changed or replaced with that of someone else. There are many types of deepfakes, and they can be created with the use of many different methods. One type of deepfake is a face-swap. These involve switching one face with another face. One method involving audio is known as lip syncing. Lip syncing is a method where the mouth movements used by a person in a video are altered to appear as if they are saying different, altered words. Another method is face synthesis, which is creating a fake image of a face by altering features of a real face image.

### 2.2. Neural Networks

Common tools used in many recent deepfake detection methods are neural networks. A neural network is a type of machine learning that is based on the human brain, mimicking the way in which neurons communicate with each other [1]. They are made up of layers of artificial neurons that are connected to each other. The neurons can be activated, which sends data to the subsequent layer of the network. Neural networks can be used in deepfake detection because they can be trained. A network is fed information such as a deepfake dataset and is able to learn from the information it takes in and predict deepfakes. As training goes on, a network becomes more accurate the more training it goes through. One type of neural network that is used often in deepfake detection is a convolutional neural network, or CNN. A CNN can take an image as input and determine the importance of different features or artifacts in the image [2]. It can then reduce the pixels of the image to create a simpler form that is easier to process.

## 3. LITERATURE REVIEW

### 3.1. Neural Network-based Deepfake Detection Methods

Deepfake detection methods involving neural networks are the most common today, and researchers have developed many that expand on the basic usages of these tools.

#### 3.1.1. Audio-focused

One form of deepfake is audio deepfakes. Researchers in [3] analyzed the relationship between audio and visual components of deepfake videos. They created a sync-stream model in which the audio and video are first fused and then analyzed by a neural network. The synced videos were trained on a CNN and tested on existing datasets. Researchers found that their sync-stream model performed better than videos where audio and visual components were not synced or were synced later in the analysis process. The sync-stream model reached a maximum accuracy of 97.62%.

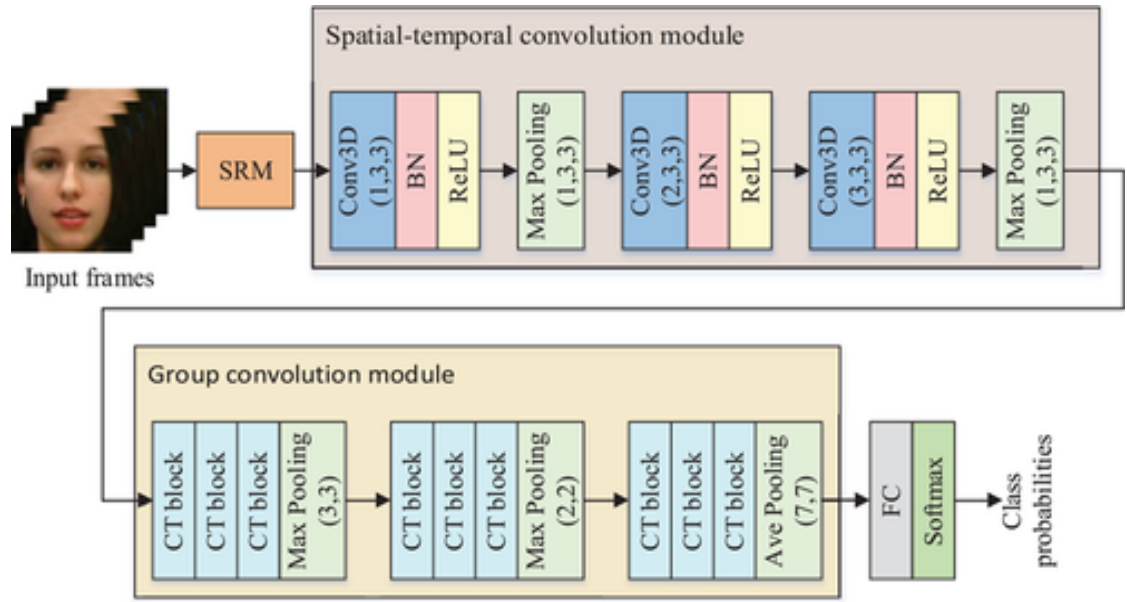

Figure 1. Proposed Architecture (Liu et al 2021)

In another study [4], researchers created a new neural network with a focus on speed and low computational requirements. Its purpose was to detect audio deepfakes. The proposed network, SpecRNet, processes two-dimensional spectrogram information of potential deepfake audios. The method performed well in terms of speed, 40% faster than LCNN, one of the fastest known methods. It also uses more trainable parameters than compared methods. Despite this, it obtained slightly lower AUC values with a maximum of 99.9322.

### 3.1.2. Identity-focused

Identity-focused deepfake detection involves using neural networks to determine and utilize the identity featured in a deepfake. A new method proposed in [5] called ID-Reveal uses a neural network to learn human facial features used in speech. The technique uses real videos to learn the biometrical traits of real faces [6]. It can then use this knowledge to recognize inconsistencies in potential deepfakes [7]. This method is strong in terms of generalizability and robustness, maintaining an accuracy of 81.8% when tested on low quality videos when compared methods performed near to 50%.

Another method [8] uses the entire face of a known identity to determine if a potential deepfake matches the identity. An algorithm called OuterFace was created that uses more of the entire image than most artifact-driven methods. It also outperforms them when used on known faces, with accuracy reaching 98.24%. One limitation is that the identity of the face must be known for detection [9]. This research also presents a new deepfake and real video dataset, Vox-DeepFake.

### 3.1.3. Feature-focused

Other neural network-based methods focus on individual facial features in detection. One example of this [10] involves a CNN used to collect facial features from each frame of a video, and an RNN (recurrent neural network) to detect inconsistencies in the features created by face swapping. This method was found to have 97% accuracy in detecting a deepfake within two seconds of video.

Other researchers in [11] recognized that traditional 3-dimensional CNNs take up a large amount of memory and storage. They created a lightweight architecture that extracts features from an image with fewer parameters. The module takes four video frames as an input and outputs features that can then be analyzed by a neural network. The results showed that they were able to reduce the number of parameters while still maintaining up to 99.83% accuracy, making it more effective for deepfake detection.

One study [12] used a convolutional vision transformer made up of a CNN and a ViT (vision transformer). The CNN was used to extract and learn features, and the ViT took them in and categorized them using an attention mechanism. This model reached 91.5% accuracy, with the limitation that the model struggles to classify and remove non-face images during testing.

Another method [13] known as ADD (attention-based deepfake detection) uses two components, face close-up and face shut-off data augmentation methods. The framework starts by localizing the face, followed by extracting the localized features, and then using feature detection to conduct face close-up and face shut-off data enhancement, altering the image to then be more easily classified by a neural network. ADD outperformed several other methods when combined with various neural networks, with up to 98.3% accuracy.

Another approach [14] is an architecture with a CNN base that involves a convolutional attention mechanism. The CNN will extract facial features, which will be put through channel and spatial attention, which analyze the inter-channel and inter-spatial relationships between the features, respectively, in order to classify the image. The testing of this method was made to mimic court cases involving deepfakes. This method achieved 91.67% accuracy when tested on the same types of deepfakes it was trained on but experienced a large drop in performance on other datasets, with a maximum of 69.84%.

One method [15] used a ST-DDL (spatial-temporal deepfake detection and localization) network to explore both temporal and spatial features simultaneously to detect fake regions of a face. The network includes an algorithm known as AMM (anchor-mesh motion) that extracts the temporal features and models the movements of micro-expressions on a face. This allows the model to better capture small features. The network then fuses the spatial and temporal features for detection. The method was tested on video-level and pixel-level criteria and performed well at the video level on all three datasets used for testing, with an AUC of .912, and the ability to learn more distinctive features.

Another method that uses both spatial and temporal features [16] used a YOLO (you only look once) algorithm with a local binary pattern histogram. The YOLO algorithm detects a face in a video frame or image. The spatial features are extracted by an algorithm. The local binary pattern histogram takes the spatial features as input and extracts temporal features. The method is faster than others tested, and results showed accuracy of up to 98.12%.

A group in [17] recognized the limitations of current deepfake detection methods in that many datasets are oversampled, creating overfitted models. The number of faces used is too small. They created a method, Face-Cutout, that cuts out part of an image of a face in order to only feed relevant information to a CNN for detection. Face-Cutout uses facial features to create multiple shapes. The shape with the largest area is chosen as a cutout from the face. When tested on two CNNs, results showed that face-cutout improves results over the baseline and Random-Erase face cutouts by reducing loss and achieving an AUC of 95.66.

### 3.1.4. Emotion Detection

One study [18] used a LSTM (long short-term memory network) to predict the emotion of a face in a potential deepfake. The predictions were made based on extracted LLDs (low level descriptors) from the visual components of a face and audio clues. The valence-arousal model of emotion was used to classify the emotions, a scale of valence (positivity or negativity) and arousal (level of excitement). The LSTM was able to determine whether the found emotion was present in both the visual and audio components of a video, and whether these results were likely for a real video. Results showed that the network had high accuracy in detecting deepfakes with the use of their audios, showing that deepfake videos are much better at replicating human emotions visually than they are audibly. This result has been corroborated by similar studies [19]. This method was also most accurate with longer video samples, up to 99.5%.

### 3.1.5. Eye Blink Detection

A model known as DeepVision [20] was created to analyze eye blinking patterns for the purpose of deepfake detection. Human blink patterns vary greatly with time of day, biological factors such as age and gender, cognitive activity, physical conditions, and level of information processing [21]. The model, using GANs, takes parameters such as these, tracks the blink patterns of a video, and determines authenticity based on how they match up with the parameters [22, 23]. Results showed 87.5% accuracy, with the limitation that other factors not considered could also affect eye blinking.

### 3.1.6. Forensic Trace Detection

Researchers in [24] developed a method to detect forensic traces in deepfake images that are created during the generation process. The method uses an EM (Expectation Maximization) algorithm to extract specific features and recreate the deepfake generation process. The EM was used to test deepfake images generated by six GANs (generative adversarial networks). Results showed at least 87% accuracy in all tests.

Another study [25] recognized that deepfake creation algorithms can only make videos of limited resolution or size, so the deepfake faces must often be warped to fit a real face. This process leaves traces in the deepfake. They captured these with CNNs by comparing the faces with their surrounding areas. They also aligned the faces with multiple sizes to simulate different resolutions in training. This method achieved up to 99.9% accuracy when tested on different existing algorithms.

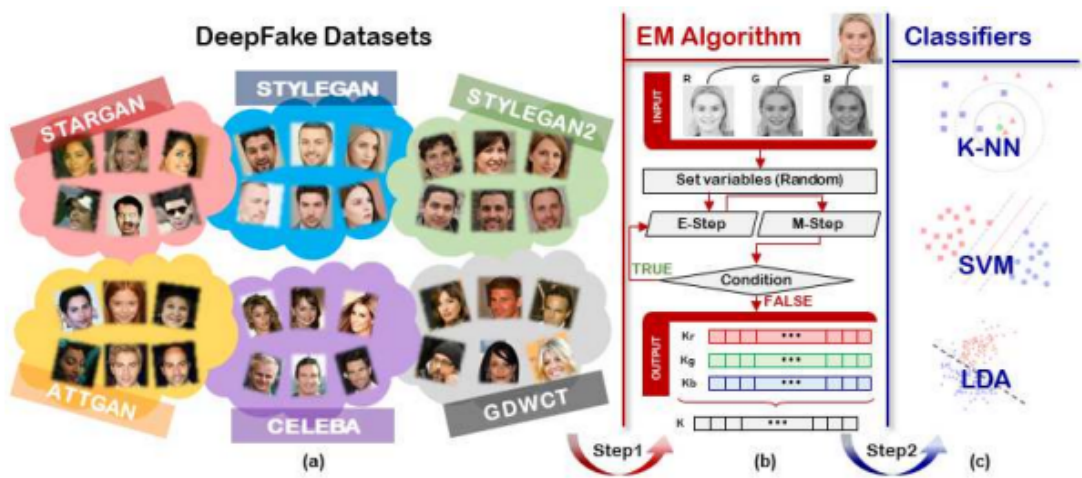

Figure 2. Deepfake Detection Pipeline (Guarnera et al 2020)

### 3.2. Corrupted Videos

In [26], researchers simulated data corruption techniques in order to test a popular deepfake neural network, EfficientNet B7, on corrupted videos. They created a video corruption pipeline by applying transformation techniques. Both real and fake videos were corrupted and tested. The videos had no audio, and the deepfakes were created with face swaps. The types of corruption tested included 720p, 480p, and 240p videos, data moshing, and bitrate.

There was a large amount of variation in results with different types of corruption, with accuracy dropping as low as 58.6% with certain corruptions and reaching a maximum of 95.9%.

### 3.2.1. Satellite Images

Researchers in [27] tested four different CNNs on both real and deepfake satellite images. They also tested seven handcrafted methods that used spatial, histogram, and frequency features on the same images. All methods were tested on real images, images distorted with JPEG compression, and images distorted with Gaussian noise. These distortions were only applied during testing. The CNNs performed much better on the real satellite images (an F1 of up to .99) but results varied with the distorted images. CNNs still achieved the highest F1 scores on both types of distorted images. With JPEG compression, the maximum was .991, and with Gaussian noise, .911. All methods performed the worst on the images distorted with Gaussian noise.

### 3.2.2. Other

A more general use of neural networks [28] involved the case where networks are trained on deepfakes generated by a different algorithm than the deepfakes they are tested on. The method extracts features from a backbone network, fine-tunes it, and outputs a score for how real an image is. Results showed that the model outperformed all but one it was tested against on average accuracy, with 75.57%.

## 3.3. Non-Neural Network-based Deepfake Detection Methods

There are deepfake detection methods that utilize tools other than neural networks, and even more that test neural networks against alternative methods.

### 3.3.1. Human Deepfake Detection

One study [29] used online testing to determine human deepfake detection performance and compare it to that of a CNN. Two experiments were conducted, one where humans were shown a real version and a fake version of the same video and asked to decide which was real and which was a deepfake with no prior knowledge, and one where they were shown a single video and asked to decide if it was real or fake. In the second experiment, the participants would be told whether the network thought the video was real or fake after they answered. They would then get a chance to change their answer if desired. In both experiments, the humans were also shown random emotional triggers before they could answer. Results showed that in the first experiment, 82% of people performed better than the network, which performed at 65% accuracy. The accuracy decreased when the videos were viewed for a longer period. In the second experiment, less people outperformed the model, but the humans still performed better on average. The emotional triggers had very little impact on results. One potential issue with this method is human bias and the use of self-selection.

Other researchers conducted a similar experiment [30] in which they set up a website that allowed humans to attempt deepfake detection and compete with two algorithms based on Xception and EfficientNet B4. Videos used were split into five categories ranging from 'very easy' to detect, to 'very difficult'. When tested on 60 videos, humans were better at identifying real videos, at 82.2%, and their overall results were more consistent. Humans performed much better than the networks on the 'easier' categories of videos, at 72.1% correct, while the opposite was true for the 'difficult' categories, when people scored only 24.5% accuracy. Both groups had similar accuracies overall, but their strengths were very different. This research shows how far deepfakes have advanced, to the point where most people cannot identify good quality deepfakes.

Another example of human deepfake detection [31] is an experiment researchers set up in a game format that allowed humans to compete against an LSTM to determine if an audio was fake. People were given one audio at a time and were prompted to choose if it was real or fake. They were then shown the machine's response. Each person also identified their age, native language, and level of technical knowledge. Results showed that humans and machines performed similarly in a realistic scenario, around 67%, increasing to 80% when users played over 10 rounds. Native English speakers performed better, but technical knowledge did not affect the results. Older people performed worse overall.

### 3.3.2. Hand-crafted Feature Detection

Another alternative to neural networks, explored in [32], involved using features hand-crafted by experts instead of generated by networks. Hand-crafted features are easier to interpret, have more forensic conformity, and lead to more plausible decisions. The method tests three fusion strategies and three types of feature detection (eye blinking, mouth region, and image foreground). Results showed that mouth region performed best out of the three features tested, up to 96.36%. For the fusion methods, the range of results was much smaller, and accuracy reached 97.57%.

## 3.4. Datasets

Nearly every deepfake detection study involves the use of one or more deepfake dataset for use in training and testing the chosen method. We are considering the impact of different types of datasets on the results of a study.

### 3.4.1. Gender and Race Biases in Datasets

Researchers in [33] manually labelled several popular datasets with gender labels. They then created a new dataset called GBDF that was gender-balanced and included gender labels. The researchers tested neural networks that were trained on either the popular datasets or GBDF. The results showed that male faces were correctly recognized more often than female faces when trained on the unbalanced dataset. The models trained on GBDF did not perform as well, but accuracy was more equal between genders and average accuracy still reached 91.2% compared to 97.5% when trained on EfficientNet V2-L. A limitation of this study is that the researchers had to judge whether a face was male or female, not accounting for mistakes or people who do not fit either category.

Another method [34] used existing gender and race-balance datasets to test popular neural networks and determine potential biases. The networks were trained on a popular, unbalanced dataset and then tested on two balanced sets. Results showed that there was a large difference in performance across races. The lowest accuracy was found on African faces, at up to 13.6% error rate. Some deepfakes are created with a face of one race swapped with the face of another. This is a common issue in FF++, the dataset used for training.

### 3.4.2. Multiple Face Swap Datasets

One form of deepfake seen in [35] involves a face that has been swapped with another more than once. This is known as phylogeny. Many datasets contain deepfakes that are not labelled with their creation method. This study created a dataset called DeePhy which contains deepfake videos created with one, two, or three face swaps. This dataset was tested on six algorithms. Capsule performed the best with an overall accuracy of 88.89%. However, all algorithms performed much better on real videos than deepfake videos.

## 4. DISCUSSION AND ANALYSIS

Through this research, we have analyzed and evaluated many deepfake detection methods. We have found that neural networks are very effective for deepfake detection in many cases. CNNs are the most prevalent form, producing high accuracy in many studies. While some researchers have found success in alternative methods, neural networks are the most reliable and generalizable overall. Nearly all research we have surveyed includes the use of datasets, which must be considered when comparing methods. Many methods we analyzed use the same group of datasets. Some of these are FaceForensics++, DFDC, Celeb-DF, and TIMIT-DF. Results may have been impacted by these datasets, as previously stated. It is also relevant that some methods used video deepfakes, some used photos, some audios, and others a mix of the three. A method that is effective for deepfake photos may have a differing level of success with videos.

The highly accurate results found in many methods we surveyed shows promise in solving the problem of malicious deepfake creation. Some of the methods that stood out were [8], [11], and [18]. The accuracies of these models reached 98% or higher. These results show that identity-focused detection, feature-focused detection, and emotion detection are effective uses of neural networks. Continued research in this area could minimize the real-world impact of malicious deepfakes.

This survey has provided many options for the development of new deepfake detection research. There is an opportunity to use these successful methods on different datasets to determine if high accuracy can still be achieved.

## 5. LIMITATIONS AND FUTURE WORK

One limitation of some neural network detection methods is that many are less lightweight and efficient in order to prioritize model accuracy [11]. Reducing the number of parameters and increasing efficiency will be an important focus of future work. A limitation of emotion-based deepfake detection is the variations in emotional expression between cultures and peoples. This leaves room for both error and bias when datasets do not contain people of all identities [18]. Bias is also a concern in human deepfake detection. It was found in one study [29] that humans are less accurate in detecting deepfakes of dark-skinned faces. Humans are also prone to errors that computers are not [36]. Additionally, in a controlled experiment, subjects expect to see deepfakes. However, this does not reflect humans' ability to detect deepfakes when they aren't expecting them. This study also brings up another limitation that may apply to many methods. Low-context videos are often used in experimental situations, which does not reflect the real-world deepfakes that pose a threat to society.

Going forward, we plan to apply the knowledge gained through this survey to the development of a new deepfake image detection method. Based on our results, we will be utilizing multiple neural networks to conduct this research.

## 6. CONCLUSION

We conclude through this survey that many deepfake detection methods are necessary to combat the new types of deepfakes constantly being developed. Identity-focused and feature-focused detection methods have produced some of the best results, and continuing to develop these ideas is crucial.